\documentclass[10pt,twocolumn,letterpaper]{article}

\usepackage{wacv}
\usepackage{times}
\usepackage{epsfig}
\usepackage{graphicx}
\usepackage{amsmath}
\usepackage{amssymb}

\usepackage{graphicx}
\usepackage{amsmath,amssymb} 
\usepackage{color}
\usepackage{hyperref}       
\usepackage{url}            
\usepackage{booktabs}       
\usepackage{amsfonts}       
\usepackage{nicefrac}       
\usepackage{microtype}      
\hypersetup{colorlinks=true, linkcolor=red}

\usepackage{soul}
\usepackage{multirow}

\usepackage{subcaption}     
\usepackage[utf8]{inputenc} 
\usepackage[T1]{fontenc}    
\usepackage{hyperref}       
\usepackage{url}            
\usepackage{booktabs}       
\usepackage{amsfonts}       
\usepackage{nicefrac}       
\usepackage{microtype}      
\usepackage{epsfig}
\usepackage{graphicx}
\hypersetup{colorlinks=true, linkcolor=red}

\usepackage{wrapfig}
\usepackage{algorithm}
\usepackage{algorithmicx}
\usepackage{algpseudocode}
\usepackage{tikz}
\usepackage{pgfplots}
\usepackage{amsmath}
\usepackage{sci}
\usepackage{xfrac}
\usepackage{pgfplotstable}
\usepackage{filecontents}
\usepackage{adjustbox}
\usepackage{amssymb}
\usepackage{pifont}

\usepackage[normalem]{ulem}

\usetikzlibrary{arrows,calc,shapes,snakes,positioning,matrix,arrows,decorations.pathmorphing,decorations.text}
\usepackage[sort&compress,square,comma,numbers]{natbib}
\makeatletter
\renewcommand\bibsection%
{
  \section*{\refname
    \@mkboth{\MakeUppercase{\refname}}{\MakeUppercase{\refname}}}
}
\makeatother

\wacvfinalcopy 

\def\wacvPaperID{98} 

\ifwacvfinal\fi
\begin{document}

\title{Collision Avoidance Detour for Multi-Agent Trajectory Forecasting: \\
A Solution for 2023 Waymo Open Dataset Challenge - Sim Agents}

\author{Hsu-kuang Chiu and Stephen F. Smith\\
Carnegie Mellon University\\
{\tt\small \{hsukuanc, ssmith\}@andrew.cmu.edu}
}

\maketitle
\ifwacvfinal\thispagestyle{empty}\fi

\begin{abstract}
We present our approach, Collision Avoidance Detour (CAD), which won the 3rd place award in the 2023 Waymo Open Dataset Challenge - Sim Agents, held at the 2023 CVPR Workshop on Autonomous Driving. To satisfy the motion prediction factorization requirement, we partition all the valid objects into three mutually exclusive sets: Autonomous Driving Vehicle (ADV), World-tracks-to-predict, and World-others. We use different motion models to forecast their future trajectories independently. Furthermore, we also apply collision avoidance detour resampling, additive Gaussian noise, and velocity-based heading estimation to improve the realism of our simulation result.
\end{abstract}

\vspace{-10pt}
\section{Introduction}
In the 2023 Waymo Open Dataset Challenge - Sim Agents \cite{waymo2023sim}, the predicted future trajectories of the ADV agent and other agents, denoted as World agents, need to be conditionally independent given the context information about the scene, such as the static map and agents' past trajectories. More specifically, the joint conditional probability of future trajectories of the ADV agent and the World agents needs to satisfy the following factorization equation:
\begin{equation} \label{eq:factorization}
\begin{split}
& p(S_{1:T}^{ADV}, S_{1:T}^{World} | c) = \Pi_{t=1}^T (\\
    & \pi_{ADV}(S_t^{ADV} | s_{<t}^{ADV}, s_{<t}^{World}, c)
     p(S_t^{World} | s_{<t}^{ADV}, s_{<t}^{World}, c)
    ),
\end{split}
\end{equation}
\noindent
where the time step $T$ is the prediction time horizon, $S_{1:T}^{ADV}$ represents the predicted future trajectory of the ADV agent, $S_{1:T}^{World}$ represents the predicted future trajectories of the World agents, $c$ represents the context information about the static map and agents' past trajectories, and $\pi_{ADV}$ represents the driving policy of the ADV.

\section{Method}
\begin{figure}[ht!]
\centering
\includegraphics[width=0.45\textwidth]{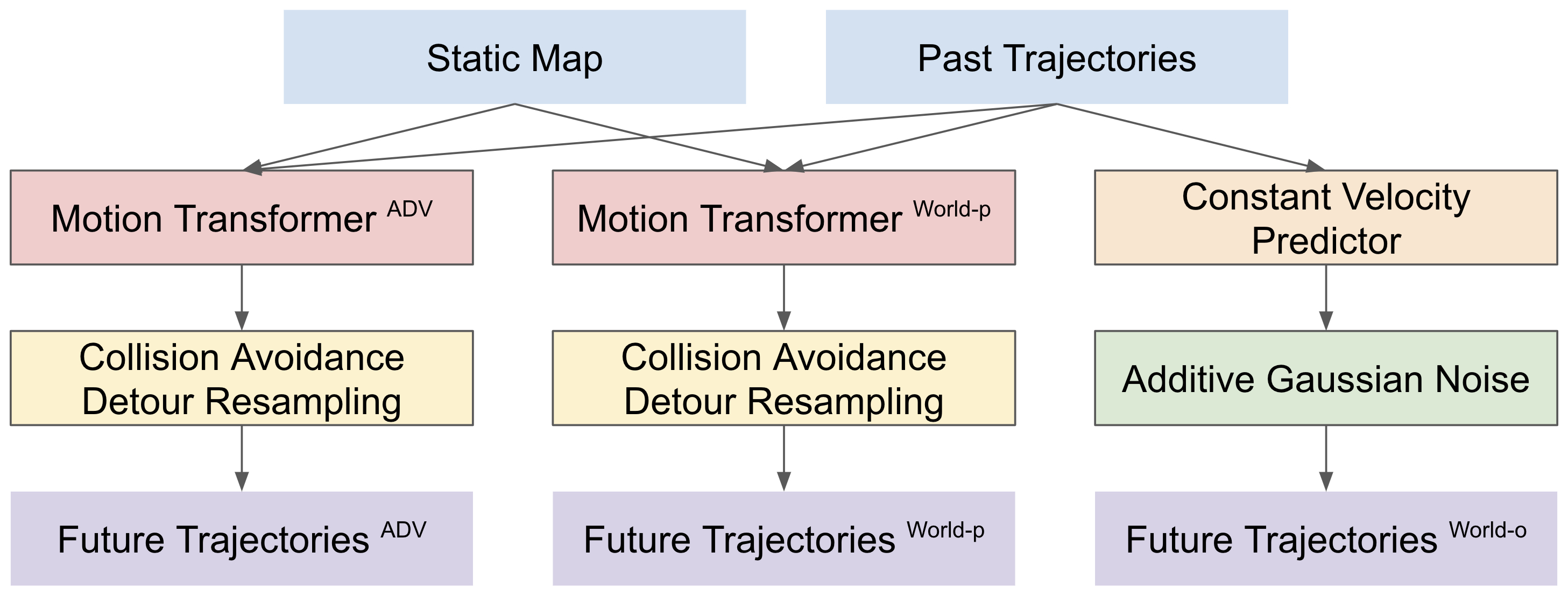}
\caption[]
        {Architecture Diagram.} 
        \label{fig:arch}
\end{figure}

Our architecture diagram can be seen in Figure \ref{fig:arch}. 

\subsection{Object Partition}
We first partition all valid objects in each scenario into three mutually exclusive groups: $ADV$, World-tracks-to-predict ($World-p$), and World-others ($World-o$). The $ADV$ group contains the single autonomous driving vehicle in each scenario of the dataset. The $World-p$ group contains other objects with the \textbf{tracks\_to\_predict} flags. The $World-o$ group contains all other valid objects.

\subsection{Motion Models}
We first train the Motion Transformer (MTR) \cite{shi2022mtr, shi2022mtra} model using the setting of Waymo Open Dataset Challenge - Motion Prediction. An MTR model takes the context information $c$ as the input and predicts $6$ future trajectories and the corresponding probability distribution for each object as the output. To satisfy the conditional independence requirement, we select the model checkpoints after training for $29$ epochs and $30$ epochs to generate the final sampled future trajectories of the object in the $ADV$ group and the objects in the $World-p$ group separately as follows:
\begin{align}
    p^{ADV}(S_{1:T}^{ADV}, S_{1:T}^{World-p} | c) &= MTR^{ADV}(c), \label{eq:mtr_adv}\\
    p^{World-p}(S_{1:T}^{World-p}, S_{1:T}^{ADV} | c) &= MTR^{World-p}(c), \label{eq:mtr_world_p}
\end{align}
\noindent
where $MTR^{ADV}$ and $MTR^{World-p}$ are two separate motion prediction models. Note that both models will predict preliminary future trajectories for objects in both $ADV$ and $World-p$ groups, but we will only keep the result $S_{1:T}^{ADV}$ from $MTR^{ADV}$, keep the result $S_{1:T}^{World-p}$ from $MTR^{World-p}$, and disregard other results in the end. Due to limited development time before the challenge submission deadline, we only train the motion transformer model using $20\%$ of training data and follow the default hyperparameter and configuration settings in \cite{shi2022mtr, shi2022mtra}.

For objects in the $World-o$ group, we use a constant velocity model with additive Gaussian noise to predict their future trajectories, the same approach as the one used in the Sim Agents challenge's tutorial:
\begin{equation}
\begin{split}
    &p^{World-o}(S_{1:T}^{World-o} | c) = \\
    &ConstantVelocityPredictor(c) +
    k \; \mathcal{N}(0, 1),
\end{split}
\end{equation}
\noindent
where $k = 0.01$ is the noise scale constant factor applied on the Gaussian noise $\mathcal{N}$ with zero mean and unit variance.

\subsection{Simulation}
During simulation, we sample future trajectories of objects in different groups independently.
For objects in the $World-o$ group, we simply sample as follows:
\begin{align}
     s_{1:T}^{World-o} \sim p^{World-o}(S_{1:T}^{World-o} | c) \label{eq:world_o}.
\end{align}

For objects in the $ADV$ and $World-p$ group,
we apply the collision avoidance detour resampling algorithm independently on the respective motion transformer outputs from equations \ref{eq:mtr_adv} and \ref{eq:mtr_world_p} without any information exchange between the two groups, which will be described in the following subsection.

\subsubsection{Collision Avoidance Detour Resampling}
The algorithm of performing collision avoidance detour resampling for a set of predicted future trajectories and the corresponding probability distribution can be seen in Algorithm \ref{alg:cad}.
The inputs include each object $i$'s 6 predicted future trajectories in an object set $O$: $S_{1:T}^i \in \mathbb{R}^{6 \times T \times 3} \; \forall \; i \in O$, and the corresponding probability distribution $p^i \in [0, 1]^{6} \; \forall \; i \in O$. The output are the sampled predicted future trajectories: $s_{1:T}^i \in \mathbb{R}^{T \times 3} \; \forall \; i \in O$.

We first sample every object's future trajectory based on the input probability distribution and then perform collision detection. If a collision happens, we simply resample all objects' future trajectories until we find collision-free joint future trajectories or we reach the maximum number of sampling trials. In our experiment, we set the maximum number of sampling
trials to $10$. We use the object center distance threshold $0.1$ meters to determine whether two objects collide with each other.

Note that we apply the collision avoidance detour resampling on the $ADV$ and the $World-p$ group independently without exchanging any information between the two groups. More specifically, to generate the final sampled future trajectory for the $ADV$, the $MTR^{ADV}$ model predicts preliminary future trajectories and probability distribution $p^{ADV}(S_{1:T}^{ADV}, S_{1:T}^{World-p} | c)$ for objects in $ADV$ groups and $World-p$ group, as shown in equation \ref{eq:mtr_adv}. Such model output is used as the input of the collision avoidance detour resampling algorithm $(S_{1:T}^i, p^i)$. And we only keep the sampled $ADV$ trajectory of this collision avoidance detour resampling process as the final sampled future trajectory of $ADV$ and disregard other objects' collision avoidance detour resampling results. All the collision detection and simulation sampling are only based on the anticipated future from the context information $c$ and the $MTR^{ADV}$ model, without using any information from the $MTR^{World-p}$ model, any predicted trajectory from the $MTR^{World-p}$ model, or any final sampled future trajectory of any object in the $World-p$ or $World-o$ groups.

Similarly, for the $World-p$ groups, the output of $MTR^{World-p}$ model $p^{World-p}(S_{1:T}^{World-p}, S_{1:T}^{ADV} | c)$, as shown in equation \ref{eq:mtr_world_p}, is used as the input of a separate collision avoidance detour resampling process. And we only keep the sampled trajectories of objects in $World-p$ from this collision avoidance detour resampling result as the final result and disregard $ADV$'s result in this collision avoidance detour resampling process. Therefore, all the collision detection and simulation sampling in this process are only based on the anticipated future from the context information $c$ and the $MTR^{World-p}$ model, without using any information from the $MTR^{ADV}$ model, any predicted trajectory from the $MTR^{ADV}$ model, or any final sampled future trajectory of $ADV$ or $World-o$ groups.

\begin{algorithm}
\caption{Collision Avoidance Detour Resampling}\label{alg:cad}
\begin{algorithmic}
\Require An object set $O$. Each object $i$'s 6 predicted future trajectories: $S_{1:T}^i \in \mathbb{R}^{6 \times T \times 3} \; \forall \; i \in O$, and the corresponding probability distribution $p^i \in [0, 1]^{6} \; \forall \; i \in O $.
\Ensure Final sampled predicted future trajectories : $s_{1:T}^i \in \mathbb{R}^{T \times 3} \; \forall \; i \in O$.\\

\State $N \gets |O|$

\For{number\_of\_trials $\gets$ 1 ... 10}
    \For{$i \gets 1 ... N$}
      \State $\hat{s}_{1:T}^i \sim (S_{1:T}^i, p^i)$
    \EndFor  
    
    \State collision $\gets$ False
    \For{$t \gets 1 ... T$}
      \For{$q \gets 1 ... N$}
        \For{$r \gets 1 ... N$}
          \If{$ q \neq r$ \textbf{and} $|\hat{s}_t^q - \hat{s}_t^r|_2 < 0.1$}
          \State collision $\gets$ True
          \EndIf
        \EndFor
      \EndFor
    \EndFor
    
    \If{collision is False $\textbf{or}$ number\_of\_trials = 10}
      \For{$i \gets 1 ... N$}
        \State $s_{1:T}^i \gets \hat{s}_{1:T}^i$
      \EndFor
      \State \textbf{return}
    \EndIf
    
\EndFor

\end{algorithmic}
\end{algorithm}

\subsection{Heading Estimation}
The aforementioned algorithm is mainly used to predict future object center positions. To predict the future heading of each object, we use the velocity-based estimation as follows:
\begin{align}
    h_t = arctan(\frac{y_t - y_{t-1}}{x_t - x_{t-1}}),
\end{align}
\noindent
where $h_t$ is the predicted object heading at time step $t$, $(x_t, y_t)$ is the predicted object center position at time step $t$, and $(x_{t-1}, y_{t-1})$ is the predicted object center position at time step $t-1$.

\subsection{Result Aggregation}
Once we finish sampling the future trajectory of each object in each group, we simply aggregate them as one final simulation rollout. And we repeat the same procedure $32$ times to generate $32$ different simulation rollouts.

\subsection{Justification}

To show that our proposed algorithm satisfies the conditional independence requirement, we present the following justification.
Given the context information $c$, our algorithm performs sampling and simulation for the objects in the three groups independently by using three independent motion models described in this section. Therefore, our algorithm satisfies the following equation:
\begin{equation}
\begin{split}
    p(S_{1:T}^{ADV}, S_{1:T}^{World} | c) = p(S_{1:T}^{ADV}, S_{1:T}^{World-p}, S_{1:T}^{World-o} | c) \\
    = p(S_{1:T}^{ADV} | c) p(S_{1:T}^{World-p} | c) p(S_{1:T}^{World-o} | c) 
\end{split}    
\end{equation}

Purely based on the formal definition of joint probability and conditional probability, we can continue to derive that:
\begin{equation}
\begin{split}
    &p(S_{1:T}^{ADV}, S_{1:T}^{World} | c) = \\
    & p(S_{1:T}^{ADV} | c) p(S_{1:T}^{World-p} | c) p(S_{1:T}^{World-o} | c) = \\
    & \Pi_{t=1}^T (p(S_t^{ADV} | s_{<t}^{ADV}, c) p(S_t^{World-p} | s_{<t}^{World-p}, c) \\
    & p(S_t^{World-o} | s_{<t}^{World-o}, c))
\end{split}
\end{equation}

Compared with the factorization equation \ref{eq:factorization}, we can see that we further factor the World agents into two groups, which is allowed based on the challenge guideline. Another difference is that our algorithm's prediction about ADV's future trajectory does not use the earlier predicted World agents' trajectories, and vice versa. And that means that our algorithm uses less available information. Similar to the algorithm used in the tutorial, using less or no information from the past predicted trajectories of agents in different groups could result in worse performance. But the result is still considered a valid submission satisfying the factorization requirement.

\section{Experiment Result}
Our algorithm's performance on the test set is shown in Table \ref{tab:evaluation_results}. Please refer to the official website of Waymo Open Dataset Challenge - Sim Agents\cite{waymo2023sim} for details about the formal definition of each evaluation metric.

\begin{table}[h!]
\footnotesize
\caption{Evaluation results on the test set of Waymo Open Dataset - Sim Agents.}
\vspace{-20pt}
\label{tab:evaluation_results}
\begin{center}
\begin{adjustbox}{width=0.35\textwidth}
\begin{tabular}{ll}
\hline
\hline
Metric Name	& Value \\
\hline
\hline
Realism meta-metric	& 0.4321 \\
Linear Speed Likelihood	& 0.3464 \\
Linear Acceleration Likelihood	& 0.2526 \\
Angular Speed Likelihood	& 0.4327 \\
Angular Acceleration Likelihood	& 0.3110 \\
Distance To Nearest Object Likelihood	& 0.3300 \\
Collision Likelihood	& 0.3114 \\
Time To Collision Likelihood	& 0.7893 \\
Distance To Road Edge Likelihood	& 0.6376 \\
Offroad Likelihood	& 0.5397 \\
minADE	& 2.3146
\vspace{-30pt}
\end{tabular}
\end{adjustbox}
\end{center}
\end{table}


{\small
\bibliographystyle{ieee}
\bibliography{refs}
}

\end{document}